\documentclass{article}
\usepackage{ijcai23}

\usepackage{times}
\usepackage{soul}
\usepackage{url}
\usepackage[hidelinks]{hyperref}
\usepackage[utf8]{inputenc}
\usepackage[small]{caption}
\usepackage{graphicx}
\usepackage{amsmath}
\usepackage{amsthm}
\usepackage{booktabs}
\usepackage{algorithm}
\usepackage{algorithmic}
\usepackage[switch]{lineno}
\usepackage{amssymb}
\usepackage{pifont}
\usepackage{multirow}

\title{GeneFormer: Learned Gene Compression using Transformer-based Context Modeling}
\author{Zhanbei Cui\textsuperscript{\rm 1,2}\thanks{This work is done when Zhanbei Cui and Yu Liao are interns at AIR, Tsinghua University},
Yu Liao \textsuperscript{\rm 1,3},
Tongda Xu\textsuperscript{\rm 1},
Yan Wang\textsuperscript{\rm 1}\thanks{Yan Wang is the corresponding author.} \\
\textsuperscript{\rm 1}Institute for AI Industry Research (AIR), Tsinghua University \\\textsuperscript{\rm 2}Beijing University of Posts and Telecommunications \\\textsuperscript{\rm 3}University of Science and Technology of China}

\begin{document}

\maketitle

\begin{abstract}
With the development of gene sequencing technology, an explosive growth of gene data has been witnessed. Thus, the storage of gene data has become an important issue. Recently, researchers begin to investigate deep learning-based gene data compression, which outperforms traditional methods based on general software like G-zip. In this paper, we propose a transformer-based gene compression method named GeneFormer. Specifically, we first introduce a modified transformer structure to eliminate the dependency of the nucleotide sequence. Then, we design a multi-level-grouping method to accelerate and improve our compression process. Experimental results on real-world datasets show that our method achieves significantly better compression ratio compared with state-of-the-art method, and the decoding speed is significantly faster than all existing learning-based gene compression methods. We will release our code on github once the paper is accepted.
\end{abstract}

\section{Introduction}

DNA is the main hereditary substance of biological object. The study on DNA is the basis of exposing biological heredity secrets \cite{1999MOLECULAR}. The order of nucleotides (or bases) in DNA (including adenine, guanine, cytosine and thymine) contains most of the genetic information. Gene sequencing technology analyzes the order of nucleotides in DNA sequence. The first method of sequencing DNA in the world was invented by Frederick Sanger~\cite{sanger1977dna}. From then on, with the rapid development of chemistry and bio-informatics, gene sequencing is becoming cheaper and faster. It is conceivable that there is a lot of DNA sequencing data generated every day. Data from the SRA~\cite{kodama2012sequence} (Sequence Read Achive) sub-database in NCBI (National Center for Biotechnology Information) shows that since 2007, the total amount of sequencing data has shown an exponential growth, doubling every 12 to 18 months, and once exceeded Moore's Law. Therefore, how to efficiently store DNA sequencing data has become an important issue.

DNA fragments carrying genetic information in DNA sequences are called genes, while others are involved in regulating the expression of genetic information or supporting DNA structure. For convenience, we call nucleotide "base" for shot, and abbreviate the four kinds of nucleotide adenine, guanine, cytosine and thymine to A, G, C and T in the rest of this paper. Similar to the problem of natural language processing, each base has relationship with bases near it. However, unlike the natural language processing with a large number of words, DNA sequence is only composed of four kinds of bases of A, G, C, T, and many fragments in the gene data appear repeatedly. It is difficult for human to understand the semantic information in the DNA sequence, for genes are always related to protein structure and DNA structures, which are nowadays not explainable for biologists.

At present, G-zip~\cite{deutsch1996rfc1952} and 7zip~\cite{pavlov20127} are commonly used compression method for DNA sequencing data. However, they are general-propose compression methods which do not make good use of the internal correlation in gene sequencing data. There should be a way to better exploit the internal characteristics of gene sequencing data to achieve better compression ratio.

On the other hand, learning based entropy estimation combined with dynamic arithmetic coding has been applied in multi-media file compression~\cite{balle2017end}. Inspired by them, several researchers have began to explore learning based method to compress gene sequencing data, including ~\cite{wang2018deepdna,cui2020compressing,sun2021complexity}. Those works all adopt LSTM~\cite{hochreiter1997long} as the main architecture to explore the correlation among bases. Those learning based approaches achieve better compression ratio than general-purpose compressors~\cite{deutsch1996rfc1952,pavlov20127} as well as previous hand-crafted gene sequence compressors including MFCompress~\cite{pinho2014mfcompress} and Genozip~\cite{lan2020genozip}.

Although learning based gene compressors are very promising and can outperform previous hand-crafted methods in compression ratio, there are still many problems waiting to be solved before they become practical for real-world applications, especially the compression latency. In this paper, to achieve a better tradeoff between compression ratio and latency, we propose a novel neural gene compressor based on the combination of transformer~\cite{vaswani2017attention} based entropy model and a multi-level-grouping method.

Our main contributions are as follows:
\begin{itemize}
    \item We are the first to introduce transformer structure for the problem of gene compression.
    \item We introduce latent array into transformer-xl~\cite{dai2019transformer} for the problem of gene compression.
    \item We design a multi-level-grouping method combining three grouping methods to improve compression ratio and reduce latency.
    \item Our model achieves state-of-the-art compression ratio, and is significantly faster than all the existing learning-based methods.
\end{itemize}

\section{Background}
DNA compression algorithms can be divided into three categories: substitution-based algorithms, statistics-based algorithms, and hybrid algorithms. 
\subsection{Substitution-based compression algorithms}
The main idea of substitution-based compression algorithms is to replace the recurring fragments in a sequence with shorter sequences and build a dictionary for them. BioCompress-I~\cite{grumbach1993compression} encodes repeats and complementary palindromes which can be precisely matched in the DNA sequence. BioCompress-II~\cite{grumbach1994new} uses order-2 arithmetic coding for non-repetitive parts on the basis of BioCompress-I. Cfact~\cite{rivals1996guaranteed} is similar to BioCompress, encoding a repeat sequence within two scans. ~\cite{chen2001compression} proposed GenCompress that has approximate matching to compress approximately duplicate fragments, which further improves the compression. Chen et al. further improved GenCompress by proposing DNACompress~\cite{chen2002dnacompress}. DNACompress first looks up all approximately duplicate sequences and then encodes each repeat in a second scan. DNAPack~\cite{behzadi2005dna} is another substitution-based compression algorithm which is based on dynamic programming. LFQC~\cite{nicolae2015lfqc} devides FASTQ files into three parts, and applies a context mixing algorithm to each of them. Genozip~\cite{lan2020genozip} is a fully functional gene compression software, which exploits known relationships between fields to improve compression.

\subsection{Statistics-based compression algorithms}
The statistics-based algorithm achieves compression by establishing probabilistic models for string representation, giving shorter coding length to more frequent characters. Loewenstern et al. proposed CDNA~\cite{loewenstern1999significantly}, which combines existence of inexact matches at various distances into a single prediction. Allison et al. proposed ARM~\cite{allison1998compression} compression algorithm later. This method first matches all repeating sequences from the original sequence and then uses the EM algorithm to calculate the model parameters. Armando et al. proposed MFCompress~\cite{pinho2014mfcompress} in 2014, which relies on multiple competing finite-context models.

Recently, some researchers have introduced deep learning into the field of gene compression. DeepDNA~\cite{wang2018deepdna} first introduced deep learning into gene compression. The author proposed a neural-network compressor containing a LSTM~\cite{hochreiter1997long} module and achieved a good result. However, his model structure is relatively simple, and there is still much room for improvement. Cui et al.~\cite{cui2020compressing} and Sun et al.~\cite{sun2021complexity} optimized it from two different perspectives. Influenced by Zhou et al.~\cite{zhou2016attention}, Cui et al.~\cite{cui2020compressing} introduces bidirectional LSTM~\cite{wollmer2010bidirectional} and attention mechanism into the algorithm, which improves the effect of the algorithm and cost more time compared to DeepDNA~\cite{wang2018deepdna}. Sun et al.~\cite{sun2021complexity} tried to introduce artificial grouping mechanism into the algorithm, and used Markov chain to reduce the storage space of initial fragment, which reduced time consumption.

\subsection{Hybird compression algorithms}
The third category combines the advantages of the former two compression methods. For example, CTW-LZ~\cite{matsumoto2000biological} calculates the compression ratio of both LZ algorithm and CTW algorithm, and selects the algorithm with lower compression ratio to compress the current base. GeNML~\cite{korodi2005efficient} uses alternative principle compression for duplicate DNA fragments, and for other sequences, it uses a context-based encoding method with a special normalized maximum likelihood discrete regression model. COMRAD~\cite{kuruppu2011iterative} compresses data by passing it for multiple times, and can be applied to large-scale optimization.

7-zip~\cite{pavlov20127}, G-zip~\cite{deutsch1996rfc1952} and bzip2~\cite{seward1996bzip2} are general-purpose data compression software. 7-zip uses Markov chain to maintain a state machine that adjusts the characteristics of the current location during compression. G-zip is based on LZ77 algorithm~\cite{ziv1977universal}, which searches for repeated strings and replaces them by the offset and the length of the matching string. bzip2 is based on the Burrows-Wheeler transformation~\cite{burrows1994block}, which achieves lossless compression by changing the order of strings to lower the entropy estimates.
\subsection{Evaluation of compression algorithms}
Generally speaking, compression ratio and time cost are two main important indicators, which are always being compared and illustrated in papers about compression algorithm. Compression ratio is the ratio of the file size before and after compression. For gene compression, it is more intuitive to use bpb (bit per base) to measure the compression ratio. Time cost refers to the time taken by the compression algorithm from start to the end of compression. An excellent compression algorithm should have lower bpb and less time cost. 

\begin{figure*}[t]
\centering
\includegraphics[width=1\textwidth]{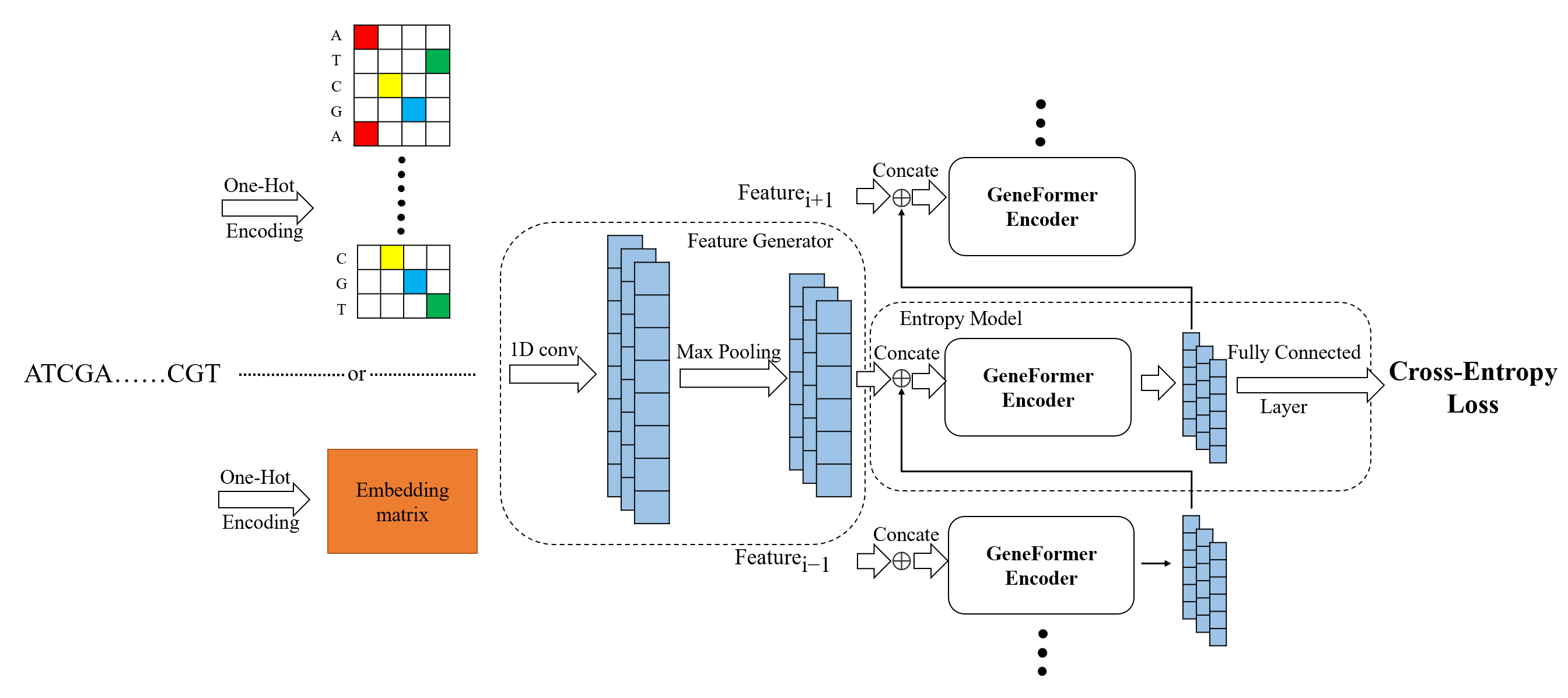} 
\caption{The structure of GeneFormer. GeneFormer is mainly composed of a feature generator and an entropy model. Bases are represented by one-hot encoding or an embedding matrix before being fed into GeneFormer. GeneFormer predicts the probability of the current base supervised by cross-entropy loss.}
\label{structure}
\end{figure*}

\section{Method}
\subsection{Overview}
\begin{figure}[t]
\flushright
\includegraphics[width=0.5\textwidth]{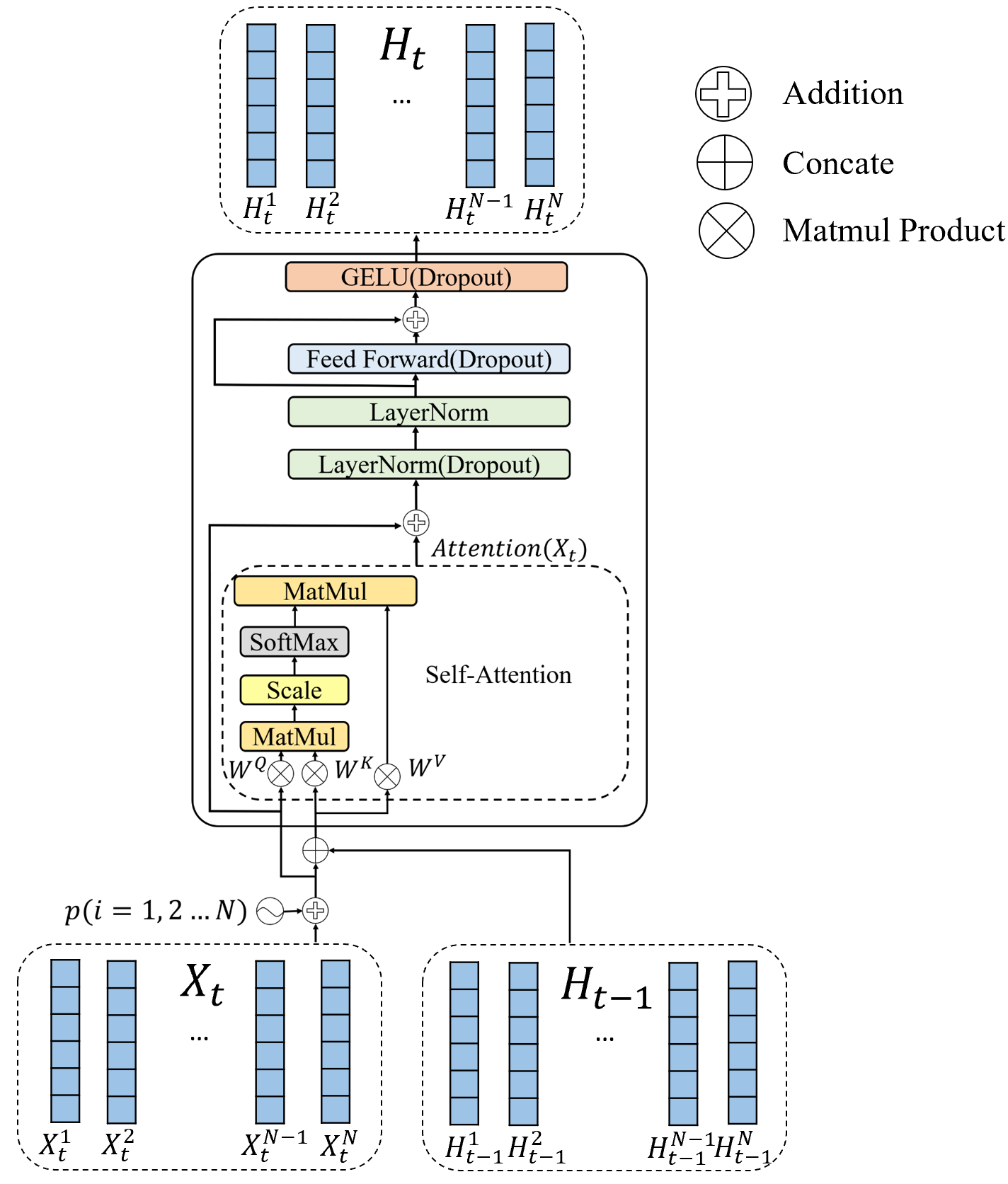} 
\caption{The inner structure of the GeneFormer encoder. For the $t^{th}$ input fragment $Z_t$, the encoder receives feature $X_{t}\in R^{N\times d_m}$ produced by feature generator and outputs latent array $H_{t}\in R^{N\times d_m}$ with the help of $H_{t-1}$. To make it easier to understand, we denote $X^i_t\in R^{d_m}$ and $H^i_t\in R^{d_m}$ as $i^{th}$ column vector of Matrix $X_t$ and $H_t$.}
\label{ENCODER}
\end{figure}

We propose GeneFormer to compress genome sequence data. GeneFormer contains two main components, a CNN-based feature generator to learn a latent representation, a transformer-based entropy model followed by a linear layer to predict the probability of the current base. We describe the structure of GeneFormer in Figure~\ref{structure}, and the detailed internal structure of the transformer-based entropy model is shown in Figure~\ref{ENCODER}. It should be noted that we set the number of attention head to 1 in Figure~\ref{ENCODER} just for convenience.

As illustrated in Figure~\ref{structure}, GeneFormer receives a sequence of bases before the current base and calculates the probability of the current base. The bases in sequence is encoded by one-hot coding or fed into an embedding layer. Both options are okay because they can reach very close scores in bpb, which are 0.0097 vs 0.0096. After GeneFormer estimates the probability, the dynamic arithmetic coding module accepts the probability and compresses the current base into binary bit stream.

\textbf{Feature generator.} We adopt a CNN-based feature generator mainly following \cite{cui2020compressing}. 1-D convolutional layer is used to extract the features in the fragment. Subsequently, a Max-Pooling layer can pick out the most important features and reduce computation cost. Detailed description is shown in Table~\ref{GeneFormer}.

\subsection{Transformer-based entropy model}
We design the architecture of GeneFormer on the basis of transformer-xl~\cite{dai2019transformer}. We select this model  because of its ability to solve the long-term dependency problem compared to the original transformer architecture~\cite{vaswani2017attention}. 

$Z$ denotes the whole gene sequence, and the $t$-th input fragment to the feature generator is written as $Z_t$. Then the output of feature generator is written as $X_{t}\in R^{N\times d_m}$ (shown in Figure~\ref{ENCODER}), which will be fed into entropy model. 

To extract the context relationship of the input feature $X_{t}$, we stack only one transformer encoder in entropy model to yield latent array $H_{t}$. $H_{t}$ is not only the output of the entropy model, but also represent the state of Geneformer. $H_{t}$ carries information of all the past segments. Transformer encoder is a time consuming module so we just use one layer of transformer-xl encoder. To generate the probability of current base, a linear layer is attached to the encoder (shown in Figure~\ref{structure}).

In transformer, dot product attention is called the vanilla attention~\cite{vaswani2017attention}. In \cite{dai2019transformer}, the authors analyse the shortcoming of vanilla transformer which is lack of ability to build long-term dependency, and propose segment-level recurrence mechanism restoring previous information to solve the problem. Following the standard notation in \cite{dai2019transformer}, we compute the matrix of attention outputs by computing the linear projection of the input $X_{t} \in R^{N\times d_m}$:
\begin{equation} 
Attention(X_{t})=\textrm{Softmax}(\frac{X_{t}W^Q(X_{t}W^K)^T}{\sqrt{d_k}})X_{t}W^V
\label{attention}
\end{equation}
where $W^Q, W^K \in R^{d_m\times d_k}$ and $W^V \in R^{d_m\times d_m}$ are learnable parameters. $d_k$ is the dimension of $W^Q$ and $W^K$.

To further adapt the transformer-xl encoder in the entropy model for gene compression, we design GeneFormer encoder. The detailed structure of GeneFormer is shown in Figure~\ref{ENCODER}. For vanilla segment-Level recurrence mechanism in transformer-xl, the length of it should be as long as possible to capture remote contextual information. However, in gene compression, we think that current base is mainly correlated to its neighbouring bases. Too long segment-level recurrence mechanism may increase the computations and slow down the convergence because of introducing too much redundant information. Therefore we cut the length to $N$, the same as $X_{t}$. As shown in the ablation experiments in Section 4, we prove the effectiveness of this modification (Segment Cut).

However, this one-size-fits-all solution is arbitrary. For the sake of over-length gene fragment and inspired by Perceiver AR~\cite{hawthorne2022general}, we introduce a latent array in the GeneFormer encoder. As transformer-xl naturally has segment-Level recurrence mechanism, we adapt it as the latent array of our encoder. After every forward propagation of input feature $X_t$, the GeneFormer stores the output $H_t$ of the encoder as latent array, and concatenates it to the next input feature $X_{t+1}$. So that the encoder can restore all the information of the segments before the current base with no extra computation. After all the improvements, the attention can be formulated as:
\begin{equation} 
{\hat{X}}_{t}=[H_{t-1},X_{t}];
\end{equation} 
\begin{equation} 
Attention(X_{t})=\textrm{Softmax}(\frac{X_{t}W^Q({\hat{X}}_{t}W^K)^T}{\sqrt{d_k}}){\hat{X}}_{t}W^V,
\end{equation} 
where $H_{t-1}$ is the GeneFormer encoder output matrix of original input $Z_{t-1}$. $[\cdot]$ means concatenating in the length dimension. 
We are the first to introduce transformer-based module into gene compression and achieve great success according to the experiments in section 4.

\textbf{Relative positional encoding.} 

Positional encoding is firstly proposed in vanilla Transformer~\cite{vaswani2017attention} because transformer encoder ignores positional information. We attach this information manually. The formula of positional encoding is
\begin{equation} 
g(i,2j)=sin(\frac{i}{t^{\frac{2j}{d_{m}}}})
\end{equation} 
\begin{equation} 
g(i,2j+1)=cos(\frac{i}{t^{\frac{2j}{d_{m}}}})
\end{equation}
where $g(.)$ means positional encoding, $i$ means the index in length, $d_{m}$ means the dimension of the feature and  $2j, 2j+1\in[0,d_{m}]$ means the index in dimension. $t$ is set to 10000 in practice.

We mainly follow the relative positional encoding in \cite{dai2019transformer}. Compared with $i$ and $j$, it puts more emphasis on $i-j$. As being formulated in Equation~\ref{attention}, before calculating $softmax(\cdot)$ and scaling, the attention score $A_{ij}$ between $i^{th}$ and $j^{th}$ token in $X_t$ can be written as:
\begin{equation} 
    A_{ij}=(X^{i}_t+g(i))W^Q((\hat{X}^{j}_t+g(j))W^K)^T 
\label{attn}
\end{equation}

Relative positional encoding reform the original encoding method by using relative position $g(i-j)$ and introducing two learnable vector parameter $u$ and $v$ to replace $g(j)$. 
If noting $W^{Q}(W^K)^{T}$ as $W$, then the Equation~\ref{attn} can be written as:
\begin{equation} 
\begin{split}
    A_{ij}=(X^{i}_{t}+u)W(\hat{X}^{j}_t)^{T}+(X^{i}_{t}+v)Wg(i-j)^{T}
\end{split}
\end{equation}

\subsection{Multi-level-grouping method}

\textbf{Fixed-length grouping (FG).} Serial decoding costs much time, inspired by \cite{sun2021complexity}, we design a fixed-length grouping method to reduce decoding latency. We set the same length for every group to limit storage space increase. 

Grouping can reduce time cost because decoder can calculate the probability of several bases from each group at the same time. However, too many numbers of groups will increase bpb (bit per base), because the initial fragment needs extra space to store. In~\cite{sun2021complexity}, the author simply divided the gene sequence into groups without careful consideration. We fix the length of bases for all the groups to trade off bpb and decoding latency. 

We further consider the acceptable extra space consumption and calculate how long each group should be to bear this extra space cost. In this way, the number of the groups depends on the length of the DNA sequence to be compressed, thus our method can fit gene sequence with various lengths.

To better reduce the storage space of the initial fragment, we use static arithmetic coding. We should carefully choose initial fragment that doesn't contain base N which means to be unknown in the dataset~\cite{mishra2010efficient} so that these sequences only contains A, G, C, T. Each base in the initial segment can be encoded in 2 bits.
Notice that we only use this grouping method during inference stage.

\textbf{Byte-grouping (BG).} \cite{mao2022fast} proposed a byte-grouping method to obtain a longer semantic dependency in general compression task. The author thinks that adjacent bytes usually contain similar information. Each byte is projected into $h/g$-dim vector space. Then they concatenate $g$ adjacent bytes into a $h$-dim vector. So that compared to the original way, this byte-grouping method can extend the context length for $g$ times. We can use this method to improve the compression process for the tensor fed into the model can be projected into 2-D space. This method will probably cause loss of information, for each embedding is smaller than before. However, the result in Section 4 shows that this method performs unexpectedly well in this task for it can reduce bpb a lot. By analysing this result, we guess that it make a dimension transformation from 1-D dimension to 2-D dimension, which just right reveal the relationship between bases.

\textbf{N-gram grouping (NG).} Gene data only contains 4 kinds of bases (A, G, C, T). As we all know that base in gene segment have combined action with its neighbor bases. For example, three consecutive bases determine one amino acid. So it may be better to group the bases before feeding them into the embedding layer. For example, we can divide the gene segment 'AGCTAT' into 'AGC' and 'TAT' if we use a 3-gram grouping method.

Besides that, this grouping method can also accelerate the compression process for it reduces the length of tensor. It also enriches the variety of embedding, for N-gram grouping method has $4^N$ kinds of embedding.

\begin{figure}[t]
\centering
\includegraphics[width=0.5\textwidth]{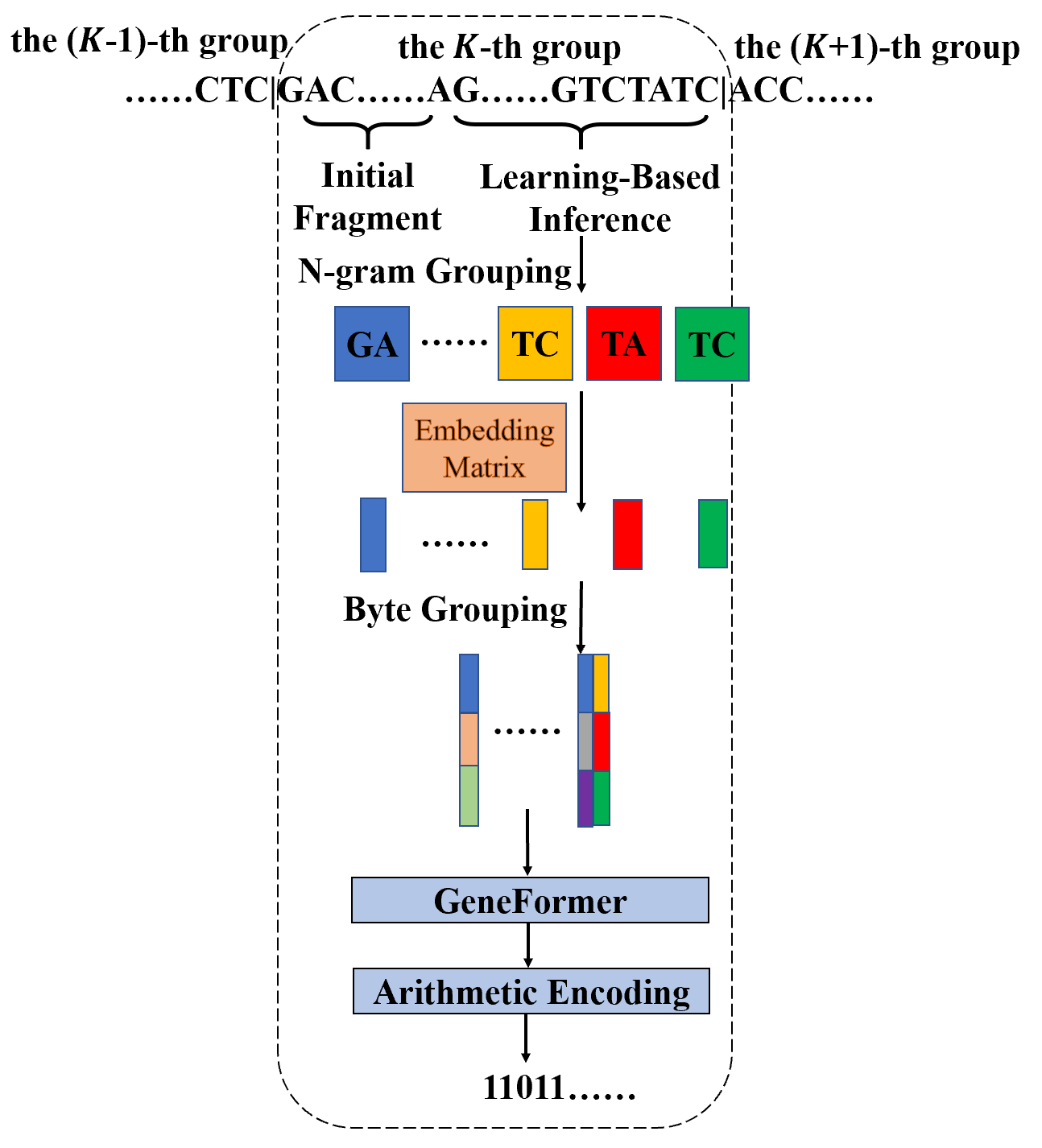} 
\caption{Illustration of the encoding and decoding process in practical application. This figure shows how GeneFormer is combined with our multi-level-grouping method. During compression in the inference stage, we cut the gene sequence into several continuous segments in fixed-length grouping method. We show the compression process of the $k$-th segment. 2-gram grouping method is used. Every two bases are regarded as a 'word' and are embedded together, thus can reduce the length of the tensor. Finally before being fed into GeneFormer, the embeddings are reshaped as a 2-D tensor using byte-grouping method. At last, the tensor is fed into Geneformer to get the probability thus the current base can be compressed by arithmetic coding.}
\label{fig3}
\end{figure}

We combined the above 3 kinds of grouping methods to improve our compression and decompression process and call the combination multi-level-grouping method, for we group the data in several places of the whole process. The latter grouping method is based on the former method. The detail of the multi-level-grouping method is shown in Figure~\ref{fig3}.

\subsection{Dynamic arithmetic encoding}
The arithmetic coding is a kind of lossless compression method approaching the entropy bound of known distributions. We encode the gene sequence using arithmetic coding as described in Algorithm~\ref{alg:ac-algorithm}.

\begin{algorithm}[tb]
\caption{Dynamic arithmetic encoding}
\label{alg:ac-algorithm}
\textbf{Input}: $p_i$ means the probability of i-th base to be A, G, C, T, N.\\
\textbf{Parameter}: $P^{l}_i$, $P^{r}_i$ means the left and right boundary of the compressed data containing the i-th base and bases before it, $p^{l}_i$ means the sum of the probability in the left of the true i-th base, $p^{r}_i$ means the sum of the probability in the right of the true i-th base. L means the length of the data to be compressed.\\
\textbf{Output}: Compressed data
\begin{algorithmic}[1] 
\STATE Arrange A, G, C, T, N in order.
\STATE Let $P^{l}_{0}=0, P^{r}_{0}=0, i=0$.
\WHILE{$i<L$}
\STATE Calculate $p^{l}_i$ and $p^{r}_i$ by $p_i$.
\STATE $P^{l}_{i+1}=P^{l}_i+p^{l}_i*(P^{r}_{i}-P^{l}_{i})$
\STATE $P^{r}_{i+1}=P^{r}_i-p^{r}_i*(P^{r}_{i}-P^{l}_{i})$
\STATE $i=i+1$
\ENDWHILE
\STATE \textbf{return} $P^{l}_{L}, P^{r}_{L}$
\end{algorithmic}
\end{algorithm}

After getting $P^{l}_{L}$ and $P^{r}_{L}$, we can easily choose a floating-point number in $[P^{l}_{L}, P^{r}_{L}]$ to represent current encoding state. For decoding, the process is just the inverse of the algorithm above.

\subsection{Encoding and decoding process}
The whole encoding and decoding process is illustrated in Figure~\ref{fig3}. During encoding, a sequence of bases before the current base is embedded and fed into GeneFormer to get the probability estimation of current base. Then the current base can be compressed by arithmetic coding. 
Bases in each fixed-length group are fed into GeneFormer simultaneously. For decoding, a sequences of bases which are already decoded before the current base are sent into the network to predict the probability of current base. Like encoding process, bases from all the groups with same relative position can be decoded simultaneously.
\begin{table*}[]
\centering
\begin{tabular}{ccccc}
\toprule
Method   Name                 & Human (bpb)      & Human (Time Cost) & Fish (bpb)       & Fish (Time Cost)            \\ \midrule
G-zip~\cite{deutsch1996rfc1952}                          & 1.4232          & 0m0.15s         & 2.4460          & 0m0.62s         \\
7zip~\cite{pavlov20127}                          & 0.0910          & 0m0.38s         & 1.2371          & 0m3.43s         \\
bzip2~\cite{seward1996bzip2}                         & 0.3558          & 0m0.22s         & 2.0303          & 0m0.44s         \\
MFCompress~\cite{pinho2014mfcompress}                   & 1.5572          & 0m1.56s         & 1.4119          & 0m5.44s         \\
Genozip~\cite{lan2020genozip}                       & 0.1360          & 0m2.36s         & 1.2869          & 0m1.87s         \\ \midrule
DeepDNA~\cite{wang2018deepdna}                       & 0.0336          & 49m53s          & 1.3591          & 133m14s         \\
DNA-BiLSTM~\cite{cui2020compressing}                & 0.0145          & 56m47s          & 0.7036          & 146m34s         \\
GeneFormer(ours)                    & 0.0097 & 82m18s          & 0.6587 & 232m57s         \\
GeneFormer+Byte-grouping (ours)                    & \textbf{0.0075} & 84m34s          & \textbf{0.4085} & 229m07s         \\
GeneFormer+Multi-level-grouping (ours)           & 0.0010          & \textbf{11m24s} & 0.4794          & \textbf{30m09s} \\ \midrule
DeepDNA (hybrid dataset)        & 0.0701          & -               & 1.1999          & -               \\
DNA-BiLSTM (hybrid dataset) & 0.0360          & -               & 1.0550          & -               \\
GeneFormer (hybrid dataset)     & 0.0215          & -               & 0.8634          & -               \\ \bottomrule
\end{tabular}
\caption{Compression performance of various methods.}  
\label{performance}
\end{table*}

\section{Experiment}
\subsection{Experimental Settings}
\textbf{Datasets}. GeneFormer is a learning-based method. We use two datasets in the experiment to prove the superiority of our method. Following DeepDNA~\cite{wang2018deepdna}, the first dataset contains 1000 human complete mitochondrial sequences with 1,691,705 bases which is downloaded from MITOMAP~\cite{brandon2005mitomap} database. Following ~\cite{wang2018deepdna}, the second dataset contains 2851 mitochondrial sequences of various fish species which is downloaded from National Center for Biotechnology Information~\cite{wheeler2007database} containing 4,564,117 bases. For each dataset, we use 70\% of the dataset for training, 20\% for validation and 10\% for testing. Especially, there are some 'N' bases in the fish dataset meaning to be uncertain according to \cite{mishra2010efficient}. For each current base to be encoded or decoded, we gather 64 bases before it and feed the sequence into GeneFormer.

\textbf{Other settings}. We do all the experiments on the same desktop with a RTX 3090Ti in a Ubuntu system. We use PyTorch to implement our method. To allow fair comparison, we mostly follow the experimental settings in~\cite{wang2018deepdna} and~\cite{cui2020compressing}. We use RMSProp optimizer. The batch size is set to 64 and the initial learning rate is set to 0.001. We do not add any learning rate scheduler, data augmentation or something else because we focus on the network architectures rather than possible improvements brought by training strategy. Almost all the hyperparameters during training are the same as the above two paper. We run all the experiments for 100 epochs. Our optimization target is a cross-entropy loss. All the GeneFormer experiments have loaded pre-trained model provided by hugging face~\cite{wolf-etal-2020-transformers}. 

For the hybrid dataset experiments, we want to explore the generalization ability of our model, so we mix the two datasets and force the model to learn both the datasets at the same time. We set the batch size of each dataset to 64 and concatenate them in the batch dimension. We set the learning rate to 0.002 for these experiments. During training stage, we calculate the bpb score in validation dataset after every training epoch. We show the best results of all the methods in Table~\ref{performance} which have the best bpb score during validations in the training stage.

\textbf{Comparisons}. Based on the datasets mentioned above, we compare GeneFormer with multiple methods including (a) G-zip, 7-zip and bzip2, which are popular compression methods, (b) MFComporess~\cite{pinho2014mfcompress}, a traditional and state-of-the-art method in statistics-based compression algorithms. Genozip~\cite{lan2020genozip},  a state-of-the-art method in substitution-based compression algorithms. (c) DeepDNA~\cite{wang2018deepdna}, the first end-to-end algorithm for DNA compression. The method using Bi-LSTM and attention module~\cite{cui2020compressing}, which is state-of-the-art in compression ratio.

\begin{table}[]
\begin{tabular}{cccll}
\cline{1-3}
\multicolumn{1}{c}{}                                    & Group Name         & Output Size &  &  \\ \cline{1-3}
\multicolumn{1}{c}{\multirow{5}*{Fearure Generator}}  & 1DConv             & 768 x 41    &  &  \\
\multicolumn{1}{c}{}                                    & Relu               & 768 x 41    &  &  \\ 
\multicolumn{1}{c}{}                                    & 1DMaxPooling       & 768 x 13    &  &  \\ 
\multicolumn{1}{c}{}                                    & BatchNormalization & 768 x 13    &  &  \\
\multicolumn{1}{c}{}                                    & Dropout(p=0.1)     & 768 x 13    &  &  \\ \cline{1-3}
\multicolumn{1}{c}{} & RelativeAttention  & 768 x 13    &  &  \\ 
\multicolumn{1}{c}{}                                    & LayerNorm          & 768 x 13    &  &  \\ 
\multicolumn{1}{c}{}                                    & Dropout(p=0.1)     & 768 x 13    &  &  \\ 
\multicolumn{1}{c}{}                                    & LayerNorm          & 768 x 13    &  &  \\ 
\multicolumn{1}{c}{\multirow{9}*[+9.5ex]{GeneFormer Encoder}} & Linear             & 3072 x 13   &  &  \\ 
\multicolumn{1}{c}{}                                    & Linear             & 768 x 13    &  &  \\ 
\multicolumn{1}{c}{}                                    & Dropout(p=0.1)     & 768 x 13    &  &  \\
\multicolumn{1}{c}{}                                    & GELUActivation     & 768 x 13    &  &  \\
\multicolumn{1}{c}{}                                    & Dropout(p=0.1)     & 768 x 13    &  &  \\ \cline{1-3}
\multicolumn{2}{c}{Linear}                                                   & 1 x 4       &  &  \\ \cline{1-3}
\end{tabular}
\caption{GeneFormer model structure.}
\label{GeneFormer}
\end{table}

\subsection{Configurations of GeneFormer}
Table~\ref{GeneFormer} shows the detailed configuration of GeneFormer. GeneFormer is mainly composed of two parts, feature generator and entropy model (GeneFormer Encoder). Feature generator mainly contains a 1-D convolutional layer and a max-pooling layer to extract and select the most obvious features. BatchNormalization layer and dropout layer are adopted to avoid overfitting. For entropy model, the most important components are self-attention module and feed-forward module containing two linear layers. Besides, layernorm layer and dropout layer are applied. In addition, we use GELU~\cite{hendrycks2016gaussian} as the activation function to avoid vanishing gradient.

In the testing stage, we set the length of each group to 213 thousand. We have to additionally store the starting 64 bases (the initial fragment) for each group, which will cause 0.0006 bpb increase on average.

\begin{table}[]
\centering
\begin{tabular}{ccc}
\toprule
Latent Array & Segment Cut & Human (bpb) \\ \midrule
\ding{56}             & \ding{52}  & 0.0102     \\
\ding{52}   & \ding{56}            & 0.0122     \\
\ding{52}   & \ding{52}  & 0.0097     \\ \bottomrule
\end{tabular}%
\caption{Performance ablation study.}
\label{ablation}
\end{table}

\begin{table}[]
\centering
\begin{tabular}{lllll}\toprule
BG & NG      & FG            & bpp  & time \\ \midrule
\ding{56}  & \ding{56} & \ding{56} & 0.0097 & 82m18s \\
\ding{52}  & \ding{56} & \ding{56} & 0.0075 & 84m34s \\
\ding{52}  & \ding{52} & \ding{56} & 0.0094 & 77m17s \\
\ding{52}  & \ding{52} & \ding{52} & 0.0010 & 11m24s \\ \bottomrule
\end{tabular}
\caption{Multi-level-grouping  ablation study.}
\label{speed ablation}
\end{table}

\subsection{Compression Performance}
Table~\ref{performance} shows the result of GeneFormer compared with the baselines based on the two datasets in terms of bpb and time cost. We directly use the bpb results in DeepDNA~\cite{wang2018deepdna} and DNA-BiLSTM~\cite{cui2020compressing}. However, for fairness, we reproduce them in PyTorch to compare the time cost.

GeneFormer outperforms all other compression methods on the two datasets. The bpb of our algorithm without grouping (0.0097) is only 66.8\% of the current state-of-the-art learning-based method DNA-BiLSTM~\cite{cui2020compressing}, and only 0.7\% of the popular G-zip method. Byte-grouping method (BG) can further improve bpb, which is 0.0075, the smallest bpb among all the existing methods and saves 48.3\% compared with DNA-BiLSTM~\cite{cui2020compressing}. Fixed-length grouping (FG) and N-gram grouping (NG) can accelerate the process of encoding and decoding. 
The bpb of our algorithm with multi-level-grouping (0.0010) is still only 69.0\% of DNA-BiLSTM~\cite{cui2020compressing} (0.0145), which means our method saves 31.0\% bit rate compared with it. This amazing result mainly comes from the GeneFormer encoder which effectively captures the relationship between bases ignoring the direction. 

Though our method with grouping is still slower than traditional methods, GeneFormer outperforms all the learning-based methods in time cost. By applying the multi-level grouping method, our method can be accelerated a lot. It only need 11m24s to compress the human gene dataset, while the state-of-the-art method DNA-BiLSTM~\cite{cui2020compressing} need 84m34s to finish the process, and still get bigger bpb than ours.

For hybrid dataset experiments whose results are shown in line 10, 11, and 12, when being trained on those two datasets, GeneFormer still performs better than other learning-based methods, even better than DeepDNA trained on one of them dedicatedly, let alone traditional methods. This result proves the good generalization ability of GeneFormer.

\textbf{Ablation study.} 
As illustrated in Table~\ref{ablation}, we remove segment-level recurrence mechanism cut and latent array in order to assess their effectiveness. 
Results show that both of them are important to reducing bpb score, and the function of latent array is significant.

In Table~\ref{speed ablation}, we also explore the three methods of grouping mentioned above including byte-grouping method, fixed-length grouping and N-gram grouping. The experiments show the acceleration degree and extra expenses each method causes. Especially, byte-grouping method can improve bpb, which is analysed in Section 3.3.

\begin{table}[]
\centering
\begin{tabular}{lll}
\toprule
Method Name         & Params(M) & MACs(G) \\ \midrule
DeepDNA~\cite{wang2018deepdna}       & 4.83      & 1.57    \\
DNA-BiLSTM~\cite{cui2020compressing}          & 2.73      & 2.45    \\
GeneFormer          & 4.80      & 4.03    \\
GeneFormer+Multi-level grouping & 4.79      & 0.94   \\ \bottomrule
\end{tabular}
\caption{Flops of different methods.}
\label{flops}
\end{table}

\textbf{Params and MACs.}
As shown in Table~\ref{flops}, Our methods have slightly less number of parameters compared with DeepDNA~\cite{wang2018deepdna}.
After applying the multi-level-grouping method, the MACs of our method is less than 1G, which is apparently smaller than DeepDNA~\cite{wang2018deepdna} and DNA-BiLSTM~\cite{cui2020compressing}.

\section{Conclusion}
Facing the challenge of storing gene data, we propose GeneFormer, a transformer-based learned gene compression method. Based on transformer-xl encoder, we cut the length of segment-Level recurrence mechanism, introduce the latent array to fit DNA compression task. With GeneFormer, we can compress gene data better and achieve state-of-the-art performance in compression ratio. Also, we design a multi-level-grouping method combining 3 kinds of grouping methods to accelerate and improve this process. After applying these methods, GeneFormer still outperforms all the existing methods in compression ratio and is much faster than all the learning-based methods. We experiment on two datasets to prove the superior performance of our model. We also design ablation experiments to show the effectiveness of different components.

In the future, we intend to design better dependency modeling method to further accelerate the encoding and decoding. We will also explore the model structure and training strategy dedicated for DNA data compression. Besides, by applying grouping methods, we realize that 1-D data may be better compressed when being reshaped into 2-D data. We will keep on exploring the way to accelerate and improve the compression performance to make using deep learning-based gene compression method possible.

\bibliographystyle{named}
\bibliography{ijcai23}

\begin{thebibliography}{}

\bibitem[\protect\citeauthoryear{Allison \bgroup \em et al.\egroup
  }{1998}]{allison1998compression}
Lloyd Allison, Timothy Edgoose, and Trevor~I Dix.
\newblock Compression of strings with approximate repeats.
\newblock In {\em ISMB}, pages 8--16, 1998.

\bibitem[\protect\citeauthoryear{Ball{\'e} \bgroup \em et al.\egroup
  }{2017}]{balle2017end}
Johannes Ball{\'e}, Valero Laparra, and Eero~P Simoncelli.
\newblock End-to-end optimized image compression.
\newblock In {\em 5th International Conference on Learning Representations,
  ICLR 2017}, 2017.

\bibitem[\protect\citeauthoryear{Behzadi and Fessant}{2005}]{behzadi2005dna}
Behshad Behzadi and Fabrice~Le Fessant.
\newblock Dna compression challenge revisited: a dynamic programming approach.
\newblock In {\em Annual Symposium on Combinatorial Pattern Matching}, pages
  190--200. Springer, 2005.

\bibitem[\protect\citeauthoryear{Brandon \bgroup \em et al.\egroup
  }{2005}]{brandon2005mitomap}
Marty~C Brandon, Marie~T Lott, Kevin~Cuong Nguyen, Syawal Spolim, Shamkant~B
  Navathe, Pierre Baldi, and Douglas~C Wallace.
\newblock Mitomap: a human mitochondrial genome database—2004 update.
\newblock {\em Nucleic acids research}, 33(suppl\_1):D611--D613, 2005.

\bibitem[\protect\citeauthoryear{Burrows and Wheeler}{1994}]{burrows1994block}
Michael Burrows and David Wheeler.
\newblock A block-sorting lossless data compression algorithm.
\newblock In {\em Digital SRC Research Report}. Citeseer, 1994.

\bibitem[\protect\citeauthoryear{Chen \bgroup \em et al.\egroup
  }{2001}]{chen2001compression}
Xin Chen, Sam Kwong, and Ming Li.
\newblock A compression algorithm for dna sequences.
\newblock {\em IEEE Engineering in Medicine and biology Magazine},
  20(4):61--66, 2001.

\bibitem[\protect\citeauthoryear{Chen \bgroup \em et al.\egroup
  }{2002}]{chen2002dnacompress}
Xin Chen, Ming Li, Bin Ma, and John Tromp.
\newblock Dnacompress: fast and effective dna sequence compression.
\newblock {\em Bioinformatics}, 18(12):1696--1698, 2002.

\bibitem[\protect\citeauthoryear{Chen}{1999}]{1999MOLECULAR}
H.~Chen.
\newblock Molecular constitution of dna and its methods of determination.
\newblock {\em Physical Testing and Chemical Analysis Part B Chemical
  Analysis}, 1999.

\bibitem[\protect\citeauthoryear{Cui \bgroup \em et al.\egroup
  }{2020}]{cui2020compressing}
Wenwen Cui, Zhaoyang Yu, Zhuangzhuang Liu, Gang Wang, and Xiaoguang Liu.
\newblock Compressing genomic sequences by using deep learning.
\newblock In {\em International Conference on Artificial Neural Networks},
  pages 92--104. Springer, 2020.

\bibitem[\protect\citeauthoryear{Dai \bgroup \em et al.\egroup
  }{2019}]{dai2019transformer}
Zihang Dai, Zhilin Yang, Yiming Yang, Jaime Carbonell, Quoc~V Le, and Ruslan
  Salakhutdinov.
\newblock Transformer-xl: Attentive language models beyond a fixed-length
  context.
\newblock {\em arXiv preprint arXiv:1901.02860}, 2019.

\bibitem[\protect\citeauthoryear{Deutsch}{1996}]{deutsch1996rfc1952}
Peter Deutsch.
\newblock Rfc1952: Gzip file format specification version 4.3, 1996.

\bibitem[\protect\citeauthoryear{Grumbach and
  Tahi}{1993}]{grumbach1993compression}
St{\'e}phane Grumbach and Fariza Tahi.
\newblock Compression of dna sequences.
\newblock In {\em [Proceedings] DCC93: Data Compression Conference}, pages
  340--350. IEEE, 1993.

\bibitem[\protect\citeauthoryear{Grumbach and Tahi}{1994}]{grumbach1994new}
St{\'e}phane Grumbach and Fariza Tahi.
\newblock A new challenge for compression algorithms: genetic sequences.
\newblock {\em Information processing \& management}, 30(6):875--886, 1994.

\bibitem[\protect\citeauthoryear{Hawthorne \bgroup \em et al.\egroup
  }{2022}]{hawthorne2022general}
Curtis Hawthorne, Andrew Jaegle, C{\u{a}}t{\u{a}}lina Cangea, Sebastian
  Borgeaud, Charlie Nash, Mateusz Malinowski, Sander Dieleman, Oriol Vinyals,
  Matthew Botvinick, Ian Simon, et~al.
\newblock General-purpose, long-context autoregressive modeling with perceiver
  ar.
\newblock {\em arXiv preprint arXiv:2202.07765}, 2022.

\bibitem[\protect\citeauthoryear{Hendrycks and
  Gimpel}{2016}]{hendrycks2016gaussian}
Dan Hendrycks and Kevin Gimpel.
\newblock Gaussian error linear units (gelus).
\newblock {\em arXiv preprint arXiv:1606.08415}, 2016.

\bibitem[\protect\citeauthoryear{Hochreiter and
  Schmidhuber}{1997}]{hochreiter1997long}
Sepp Hochreiter and J{\"u}rgen Schmidhuber.
\newblock Long short-term memory.
\newblock {\em Neural computation}, 9(8):1735--1780, 1997.

\bibitem[\protect\citeauthoryear{Kodama \bgroup \em et al.\egroup
  }{2012}]{kodama2012sequence}
Yuichi Kodama, Martin Shumway, and Rasko Leinonen.
\newblock The sequence read archive: explosive growth of sequencing data.
\newblock {\em Nucleic acids research}, 40(D1):D54--D56, 2012.

\bibitem[\protect\citeauthoryear{Korodi and Tabus}{2005}]{korodi2005efficient}
Gergely Korodi and Ioan Tabus.
\newblock An efficient normalized maximum likelihood algorithm for dna sequence
  compression.
\newblock {\em ACM Transactions on Information Systems (TOIS)}, 23(1):3--34,
  2005.

\bibitem[\protect\citeauthoryear{Kuruppu \bgroup \em et al.\egroup
  }{2011}]{kuruppu2011iterative}
Shanika Kuruppu, Bryan Beresford-Smith, Thomas Conway, and Justin Zobel.
\newblock Iterative dictionary construction for compression of large dna data
  sets.
\newblock {\em IEEE/ACM transactions on Computational Biology and
  Bioinformatics}, 9(1):137--149, 2011.

\bibitem[\protect\citeauthoryear{Lan \bgroup \em et al.\egroup
  }{2020}]{lan2020genozip}
Divon Lan, Raymond Tobler, Yassine Souilmi, and Bastien Llamas.
\newblock genozip: a fast and efficient compression tool for vcf files.
\newblock {\em Bioinformatics}, 2020.

\bibitem[\protect\citeauthoryear{Loewenstern and
  Yianilos}{1999}]{loewenstern1999significantly}
David Loewenstern and Peter~N Yianilos.
\newblock Significantly lower entropy estimates for natural dna sequences.
\newblock {\em Journal of computational Biology}, 6(1):125--142, 1999.

\bibitem[\protect\citeauthoryear{Mao \bgroup \em et al.\egroup
  }{2022}]{mao2022fast}
Yu~Mao, Yufei Cui, Tei-Wei Kuo, and Chun~Jason Xue.
\newblock A fast transformer-based general-purpose lossless compressor.
\newblock {\em arXiv preprint arXiv:2203.16114}, 2022.

\bibitem[\protect\citeauthoryear{Matsumoto \bgroup \em et al.\egroup
  }{2000}]{matsumoto2000biological}
Toshiko Matsumoto, Kunihiko Sadakane, and Hiroshi Imai.
\newblock Biological sequence compression algorithms.
\newblock {\em Genome informatics}, 11:43--52, 2000.

\bibitem[\protect\citeauthoryear{Mishra \bgroup \em et al.\egroup
  }{2010}]{mishra2010efficient}
Kamta~Nath Mishra, Anupam Aaggarwal, Edries Abdelhadi, and DPC Srivastava.
\newblock An efficient horizontal and vertical method for online dna sequence
  compression.
\newblock {\em International Journal of Computer Applications}, 3(1):39--46,
  2010.

\bibitem[\protect\citeauthoryear{Nicolae \bgroup \em et al.\egroup
  }{2015}]{nicolae2015lfqc}
Marius Nicolae, Sudipta Pathak, and Sanguthevar Rajasekaran.
\newblock Lfqc: a lossless compression algorithm for fastq files.
\newblock {\em Bioinformatics}, 31(20):3276--3281, 2015.

\bibitem[\protect\citeauthoryear{Pavlov}{2012}]{pavlov20127}
Igor Pavlov.
\newblock 7-zip.
\newblock {\em URL http://www}, 2012.

\bibitem[\protect\citeauthoryear{Pinho and Pratas}{2014}]{pinho2014mfcompress}
Armando~J Pinho and Diogo Pratas.
\newblock Mfcompress: a compression tool for fasta and multi-fasta data.
\newblock {\em Bioinformatics}, 30(1):117--118, 2014.

\bibitem[\protect\citeauthoryear{Rivals \bgroup \em et al.\egroup
  }{1996}]{rivals1996guaranteed}
Eric Rivals, Jean-Paul Delahaye, Max Dauchet, et~al.
\newblock A guaranteed compression scheme for repetitive dna sequences.
\newblock In {\em Data Compression Conference}, pages 453--453. IEEE Computer
  Society, 1996.

\bibitem[\protect\citeauthoryear{Sanger \bgroup \em et al.\egroup
  }{1977}]{sanger1977dna}
Frederick Sanger, Steven Nicklen, and Alan~R Coulson.
\newblock Dna sequencing with chain-terminating inhibitors.
\newblock {\em Proceedings of the national academy of sciences},
  74(12):5463--5467, 1977.

\bibitem[\protect\citeauthoryear{Seward}{1996}]{seward1996bzip2}
Julian Seward.
\newblock bzip2 and libbzip2.
\newblock {\em avaliable at http://www. bzip. org}, 1996.

\bibitem[\protect\citeauthoryear{Sun \bgroup \em et al.\egroup
  }{2021}]{sun2021complexity}
Zhenhao Sun, Meng Wang, Shiqi Wang, and Sam Kwong.
\newblock Complexity-configurable learning-based genome compression.
\newblock In {\em 2021 Picture Coding Symposium (PCS)}, pages 1--5. IEEE, 2021.

\bibitem[\protect\citeauthoryear{Vaswani \bgroup \em et al.\egroup
  }{2017}]{vaswani2017attention}
Ashish Vaswani, Noam Shazeer, Niki Parmar, Jakob Uszkoreit, Llion Jones,
  Aidan~N Gomez, {\L}ukasz Kaiser, and Illia Polosukhin.
\newblock Attention is all you need.
\newblock {\em Advances in neural information processing systems}, 30, 2017.

\bibitem[\protect\citeauthoryear{Wang \bgroup \em et al.\egroup
  }{2018}]{wang2018deepdna}
Rongjie Wang, Yang Bai, Yan-Shuo Chu, Zhenxing Wang, Yongtian Wang, Mingrui
  Sun, Junyi Li, Tianyi Zang, and Yadong Wang.
\newblock Deepdna: A hybrid convolutional and recurrent neural network for
  compressing human mitochondrial genomes.
\newblock In {\em 2018 IEEE International Conference on Bioinformatics and
  Biomedicine (BIBM)}, pages 270--274. IEEE, 2018.

\bibitem[\protect\citeauthoryear{Wheeler \bgroup \em et al.\egroup
  }{2007}]{wheeler2007database}
David~L Wheeler, Tanya Barrett, Dennis~A Benson, Stephen~H Bryant, Kathi
  Canese, Vyacheslav Chetvernin, Deanna~M Church, Michael DiCuccio, Ron Edgar,
  Scott Federhen, et~al.
\newblock Database resources of the national center for biotechnology
  information.
\newblock {\em Nucleic acids research}, 36(suppl\_1):D13--D21, 2007.

\bibitem[\protect\citeauthoryear{Wolf \bgroup \em et al.\egroup
  }{2020}]{wolf-etal-2020-transformers}
Thomas Wolf, Lysandre Debut, Victor Sanh, Julien Chaumond, Clement Delangue,
  Anthony Moi, Pierric Cistac, Tim Rault, Rémi Louf, Morgan Funtowicz, Joe
  Davison, Sam Shleifer, Patrick von Platen, Clara Ma, Yacine Jernite, Julien
  Plu, Canwen Xu, Teven~Le Scao, Sylvain Gugger, Mariama Drame, Quentin Lhoest,
  and Alexander~M. Rush.
\newblock Transformers: State-of-the-art natural language processing.
\newblock In {\em Proceedings of the 2020 Conference on Empirical Methods in
  Natural Language Processing: System Demonstrations}, pages 38--45, Online,
  October 2020. Association for Computational Linguistics.

\bibitem[\protect\citeauthoryear{W{\"o}llmer \bgroup \em et al.\egroup
  }{2010}]{wollmer2010bidirectional}
Martin W{\"o}llmer, Florian Eyben, Alex Graves, Bj{\"o}rn Schuller, and Gerhard
  Rigoll.
\newblock Bidirectional lstm networks for context-sensitive keyword detection
  in a cognitive virtual agent framework.
\newblock {\em Cognitive Computation}, 2(3):180--190, 2010.

\bibitem[\protect\citeauthoryear{Zhou \bgroup \em et al.\egroup
  }{2016}]{zhou2016attention}
Peng Zhou, Wei Shi, Jun Tian, Zhenyu Qi, Bingchen Li, Hongwei Hao, and Bo~Xu.
\newblock Attention-based bidirectional long short-term memory networks for
  relation classification.
\newblock In {\em Proceedings of the 54th annual meeting of the association for
  computational linguistics (volume 2: Short papers)}, pages 207--212, 2016.

\bibitem[\protect\citeauthoryear{Ziv and Lempel}{1977}]{ziv1977universal}
Jacob Ziv and Abraham Lempel.
\newblock A universal algorithm for sequential data compression.
\newblock {\em IEEE Transactions on information theory}, 23(3):337--343, 1977.

\end{thebibliography}

\end{document}